\documentclass[letterpaper]{article} 
\usepackage{aaai2026}  
\usepackage{times}  
\usepackage{helvet}  
\usepackage{courier}  
\usepackage[hyphens]{url}  
\usepackage{graphicx} 
\usepackage{natbib}  
\usepackage{caption} 
\usepackage{algorithm}
\usepackage{algorithmic}

\usepackage{newfloat}
\usepackage{listings}
\DeclareCaptionStyle{ruled}{labelfont=normalfont,labelsep=colon,strut=off} 
\floatstyle{ruled}
\newfloat{listing}{tb}{lst}{}
\floatname{listing}{Listing}
\usepackage{cutwin}
\usepackage{enumitem}
\usepackage{amsmath}
\usepackage{booktabs}
\usepackage{multirow}
\usepackage{color}
\usepackage{colortbl}

\usepackage{amsfonts}       
\usepackage{nicefrac}       
\usepackage{microtype}      
\usepackage{xcolor}         
\usepackage{bbding}
\usepackage{pifont}

\title{End4: End-to-end Denoising Diffusion for Diffusion-Based Inpainting Detection}

\author{
Fei Wang\textsuperscript{\rm 1}, Xuecheng Wu\textsuperscript{\rm 2}, Zheng Zhang\textsuperscript{\rm 1}, Danlei Huang\textsuperscript{\rm 2}, Yuheng Huang\textsuperscript{\rm 3}, Bo Wang\textsuperscript{\rm 1}\thanks{Corresponding author.} 
}
\affiliations{
    \textsuperscript{\rm 1}School of Information and Communication Engineering, Dalian University of Technology, Dalian, China\\
    \textsuperscript{\rm 2}School of Computer Science and Technology, Xi'an Jiaotong University, Xi'an, China\\
    \textsuperscript{\rm 3}Nanjing Intelligent Transportation Information Co.Ltd, Nanjing, China\\
    {\tt fei\_wang@mail.dlut.edu.cn,~xuecwu@gmail.com,~bowang@dlut.edu.cn}
}

\begin{document}
\maketitle

\begin{abstract}
The powerful generative capabilities of diffusion models have significantly advanced the field of image synthesis, enhancing both full image generation and inpainting-based image editing. Despite their remarkable advancements, diffusion models also raise concerns about potential misuse for malicious purposes. However, existing approaches struggle to identify images generated by diffusion-based inpainting models, even when similar inpainted images are included in their training data. To address this challenge, we propose a novel detection method based on  \textbf{End}-to-en\textbf{d} \textbf{d}enoising \textbf{d}iffusion (\textbf{End4}). Specifically, End4 designs a denoising reconstruction model to improve the alignment degree between the latent spaces of the reconstruction and detection processes, thus reconstructing features that are more conducive to detection. Meanwhile, it leverages a Scale-aware Pyramid-like Fusion Module (SPFM) that refines local image features under the guidance of attention pyramid layers at different scales, enhancing feature discriminability. Additionally, to evaluate detection performance on inpainted images, we establish a comprehensive benchmark comprising images generated from five distinct masked regions. Extensive experiments demonstrate that our End4 effectively generalizes to unseen masking patterns and remains robust under various perturbations. Our code and dataset will be released soon.
\end{abstract}

\section{Introduction}
\label{sec:intro}

\begin{figure}[!t]
\centering
\includegraphics[width=1\linewidth]{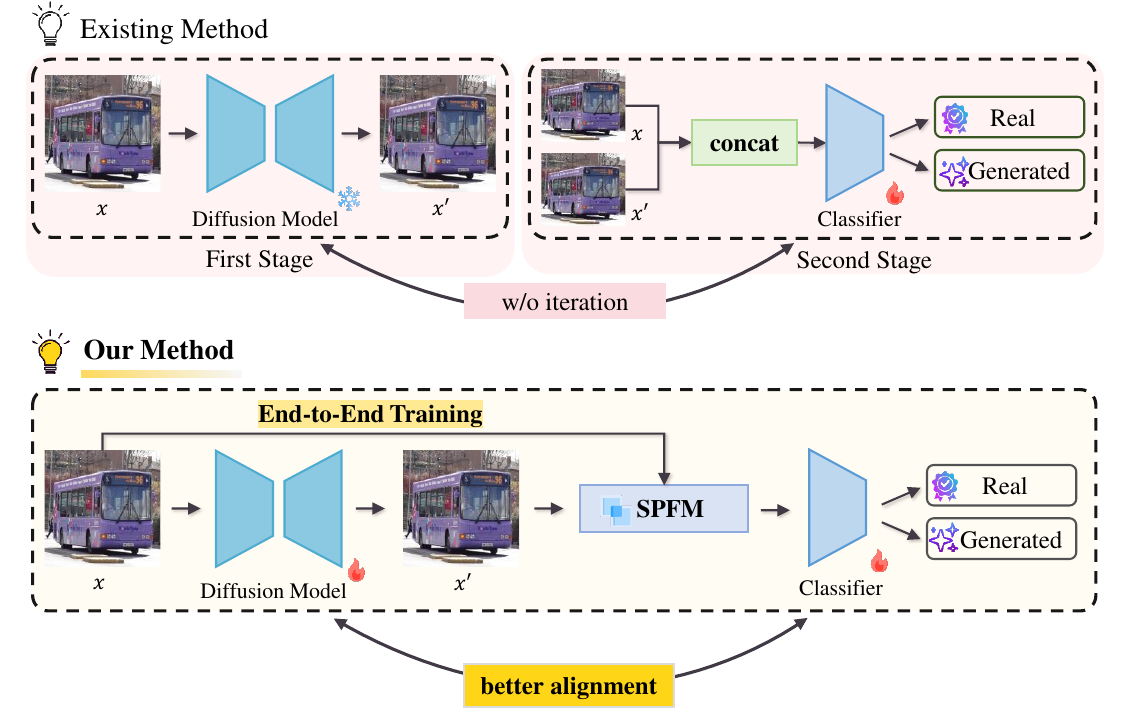}
\vspace{0.2em}
\caption{Comparison between existing reconstruction-based detection methods and ours. Existing methods operate in two stages: first, a pre-trained diffusion model generates a reconstructed image; then, the original and the reconstructed images are concatenated to train a backend classifier. Our method utilizes an updated diffusion model, enabling end-to-end learning and improving alignment of the latent spaces for the reconstruction and detection stages.}
\label{fig:intro}
\end{figure}

The emergence of diffusion models, such as Stable Diffusion (SD) \cite{rombach2022high}, DreamBooth \cite{ruiz2023dreambooth} and InstantBooth \cite{shi2024instantbooth}, has revolutionized high-fidelity and controllable image synthesis. These models have found widespread application in creative design, personalized content generation, and super-resolution imaging. However, these advancements also introduce serious security concerns. In particular, semantically guided inpainting, where specific regions of an image are modified or replaced based on textual input, can be exploited to fabricate visual misinformation, posing risks to public trust on social media platforms \cite{juefei2022countering}. Consequently, there is an urgent imperative to devise robust techniques capable of identifying diffusion-based inpainting forgeries.

Notably, existing approaches to identifying synthetic content in diffusion‐generated images primarily focus on fully generated images, overlooking more subtle, localized manipulations. For instance, current studies \cite{ojha2023towards, tan2024rethinking, chen2024learning, chen2024drct} have introduced detectors tailored to diffusion outputs, such as leveraging reconstruction errors, assuming that diffusion models can reproduce their own synthetic outputs but struggle to reconstruct real images, to distinguish generated from real images \cite{wang2023dire, chu2024fire, luo2024lare, cazenavette2024fakeinversion, ricker2024aeroblade}. However, our experiments reveal that these methods nearly collapse in detecting local inpainting images (see Section Experiments). Additionally, although Frick et al. \cite{frick2024diffseg} propose a multi‐feature segmentation framework to detect diffusion‐based inpainting generations, their evaluation is also limited in two respects: they do not assess the intrinsic authenticity of the inpainted regions, nor do they systematically study how varying mask geometries affect the detection performance.

These observations raise a critical question: \textit{why do exiting methods that perform well on fully generated images struggle to detect diffusion-based inpainting?} We hypothesize that this performance gap arises from two key limitations. First, they lack mechanisms for localized feature analysis, making them less sensitive to subtle, region-specific manipulations. Second, the reconstruction and detection are typically treated as independent processes, limiting the model’s ability to jointly optimize for forgery localization and classification. Although FIRE \cite{chu2024fire} introduces an end-to-end detection framework, it still depends on a fixed, pre-trained diffusion model and does not explicitly tailor the reconstruction module for the inpainting detection task.

Motivated by these insights, we introduce \textbf{End}-to-en\textbf{d} \textbf{d}enoising \textbf{d}iffusion (\textbf{End4}) framework for detecting diffusion-based inpainting forgeries. End4  comprises three core components: an Updated Diffusion Model (UpDM), a Scale-aware Pyramid-like Fusion Module (SPFM), and a backend classifier.
The UpDM serves as an efficient, one-step reconstruction network with two key enhancements. First, it is trained using a noise-prediction objective-minimizing the discrepancy between the ground-truth noise and the model’s prediction on generated images, thereby aligning the latent spaces of reconstruction and detection. Second, it performs denoising in one reverse diffusion step, in contrast to conventional approaches that require iterative, multi-step reconstruction, yielding substantial computational savings.
On this basis, the SPFM extracts rich, localized cues by processing both the original and reconstructed images through a cascade of attention-based pyramid layers operating at multiple spatial scales. This multi-scale design refines the capture of subtle, region-specific artifacts, while the self-attention mechanism promotes contextual feature interaction across disparate regions. In the end, the multi-scale features from both inputs undergo multi-head cross attention and are fed into the backend classifier to produce the final inpainting-forgery prediction.

Additionally, we establish a comprehensive evaluation benchmark, called InpaintingForensics, comprising images inpainted by three distinct diffusion-based models trained on the COCO dataset \cite{lin2014microsoft}. To assess robustness across diverse editing scenarios, we apply five mask types that vary in shape, size, and location. We will publicly release InpaintingForensics to establish a standardized evaluation protocol for diffusion-based inpainting detection. Extensive experiments demonstrate that, compared with existing advanced methods, End4 markedly improves detection performance in inpainting forgeries: it achieves high accuracy on images produced with unseen masks and maintains robustness against various perturbations.

In conclusion, the main contributions of this paper are three-fold:
\begin{itemize}
\item To the best of our knowledge, we are the first to update the diffusion reconstruction model for the forgery detection task.  By training with a noise-prediction objective and leveraging one-step denoising, our UpDM can better align the latent spaces of reconstruction and detection than previous approaches.
\item We identify the limitations of existing reconstruction-based methods in capturing localized inpainting artifacts and propose SPFM to address them. SPFM processes both the original and reconstructed images through cascaded attention-based pyramid layers at multiple scales, enabling fine-grained, region-aware feature extraction.
\item We evaluate End4 on our newly curated InpaintingForensics benchmark, comprising COCO-mask inpainted images with five mask types, and demonstrate substantial gains over competing methods. On the COCO-mask subset, End4 achieves a 16.34\% increase in accuracy and a 14.23\% increase in AUC, setting a new benchmark for diffusion-based inpainting detection.
\end{itemize}
\section{Related Work}
\label{sec:Related Work}

\paragraph{Generated Image Detection.} With the development of AI-based image generation technologies, numerous detection methods emerge to address the challenges posed by increasingly realistic synthetic images. 
The primary methods \cite{li2018ictu, qian2020thinking, haliassos2021lips, gao2024texture} utilize artifacts in images for deepfake detection. Some methods  \cite{shiohara2022detecting, cao2022end, le2023quality} attempt to improve the generalization of models by leveraging additional tasks. Additionally, multimodal methods \cite{yin2024improving, zou2024cross} utilize semantic information to enhance detection. Shao et al. \cite{shao2023detecting} propose HAMMER, which detects and attributes manipulated content by examining the subtle interactions between images and text.

The emergence of Generative Adversarial Networks (GANs) \cite{goodfellow2014generative} and Diffusion Models (DMs) \cite{ho2020denoising} presents new challenges for detection. Yu et al. \cite{yu2019attributing} and Marra et al. \cite{marra2019gans} extract the unique fingerprints of the GAN model from generated images to perform detection.
Ojha et al. \cite{ojha2023towards} use a pre-trained CLIP:ViT network to enable generalization between GANs and DMs. Tan et al. \cite{tan2024rethinking} introduce neighborhood pixel relationships to identify generated content through upsampling artifacts in GAN and diffusion processes.

\paragraph{Detection Based on Reconstruction Error.} The emergence of DMs drives the development of specialized detectors for identifying diffusion-generated images. Wang et al. \cite{wang2023dire} first introduce Diffusion Reconstruction Error (DIRE), which distinguishes between real and generated images by measuring reconstruction error. Luo et al. \cite{luo2024lare} propose Latent Reconstruction Error (LaRE), achieving an 8-times speed increase compared to DIRE. Ma et al. \cite{ma2023exposing} enhance detection accuracy in their SeDID method by utilizing multi-step error calculations. Cazenavette et al. \cite{cazenavette2024fakeinversion} develop FakeInversion, which detects unseen text-to-image models through text-conditioned inversion. Chen et al. \cite{chen2024drct} introduce reconstruction contrastive learning to improve generalization by generating hard samples. Ricker et al. \cite{ricker2024aeroblade} propose a training-free method by leveraging the reconstruction error of latent DMs using an autoencoder. Additionally, Chu et al. \cite{chu2024fire} observe that DMs struggle to accurately reconstruct mid-frequency information in real images, using this as a cue for detecting diffusion-generated images.

Compared to existing methods, our approach is the first to update the diffusion model during the reconstruction process rather than relying on a pre-trained model, thereby achieving better alignment of the latent spaces in the reconstruction and detection stages. Furthermore, by exploring the fusion of local and global features between the original and the reconstructed images, we construct a detector specifically targeting diffusion-based inpainting generated images.
\section{Methodology}
\label{sec:Mthods}
In this section, we first introduce the basics of denoising diffusion models. Then, we present the architecture of our method. An updated diffusion model reconstructs the original image, and then both the original and the reconstructed images are fed into a scale-aware pyramid-like fusion module. Finally, we determine whether the image is generated by diffusion-based inpainting models.

\begin{figure*}[!t]
\centering
\includegraphics[width=5.5in]{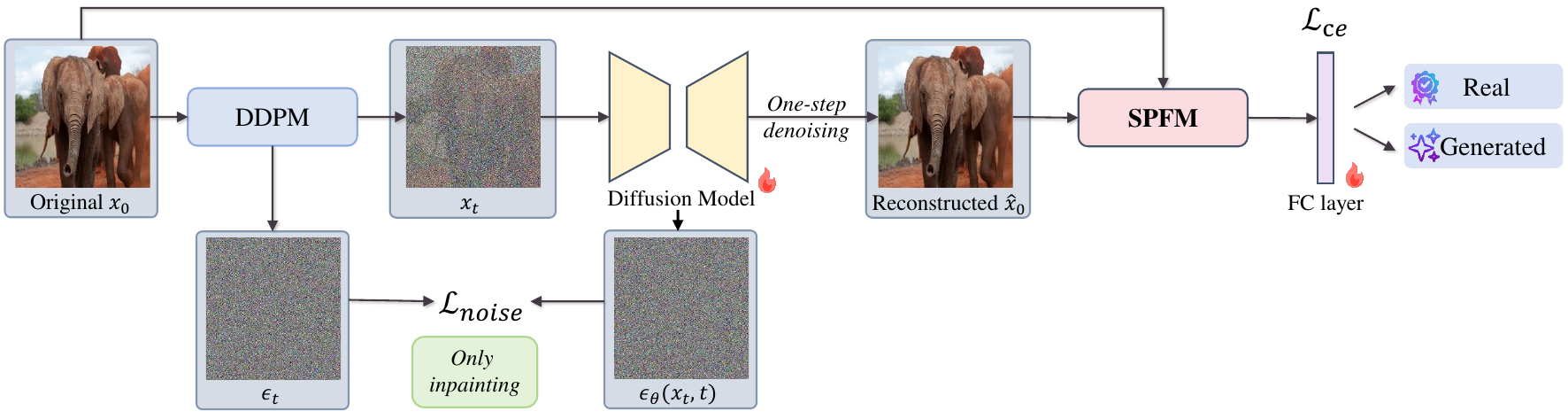}
\caption{The overview of our method. First, we use an updated diffusion model to reconstruct the original image through one-step denoising. Furthermore, we propose a Scale-aware Pyramid-like Fusion Module (SPFM) to refine the extraction of local features. Finally, the fused features are fed into a fully connected layer for real and generated classification.}
\label{pipeline}
\end{figure*}

\begin{figure*}[!t]
\centering
\includegraphics[width=5.5in]{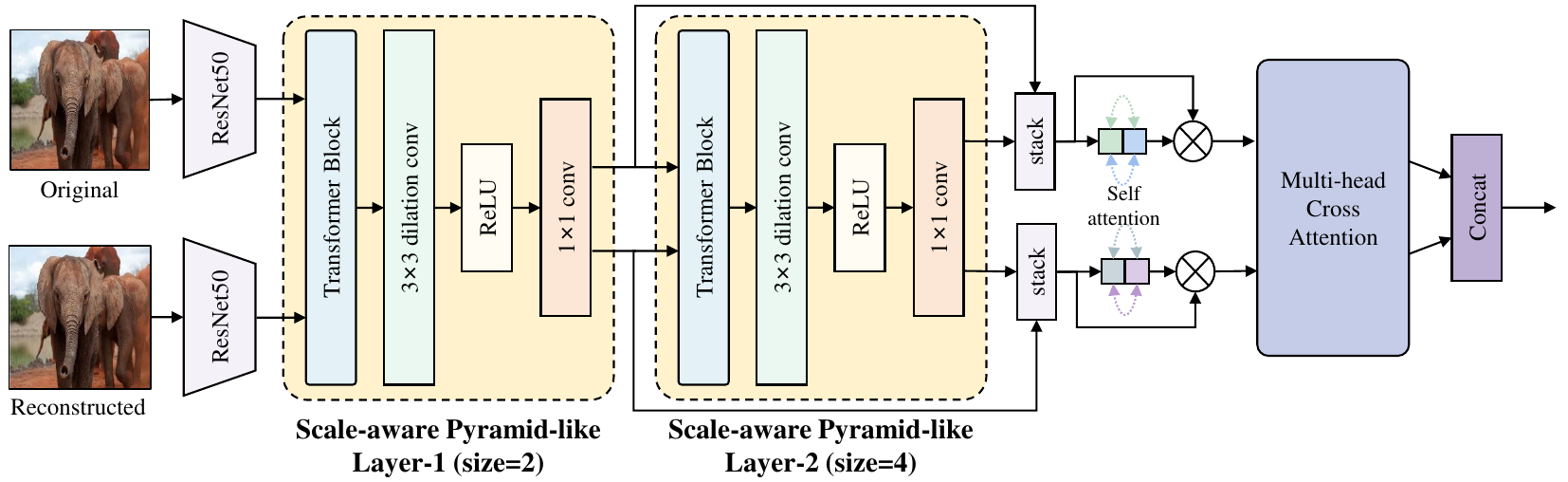}
\caption{The architecture of our proposed SPFM, which consists of two scale-aware pyramid-like layers, self attention and multi-head cross attention.}
\label{fig:SPFM}
\end{figure*}

\subsection{Preliminaries}
Existing diffusion models can be summarized in two stages: the forward process and the reverse process. In the forward process, the original image is progressively perturbed by adding Gaussian noise,
\begin{equation}
\label{eq1}
x_{t}=\sqrt{\bar{\alpha}_{t}} x_{0}+\sqrt{1-\bar{\alpha}_{t}} \epsilon_{t},
\end{equation}
where $\epsilon_{t} \sim \mathcal{N}(0, I)$ for $t = 0,..., T$. Here, $x_{t}$ represents the noisy image at time step $t$, and $x_{0}$ represents the initial image without noise. $\bar{\alpha}_{t} = \prod_{i=0}^{t} \alpha_{i}$ denotes the noise scaling factor at step $t$.

In the reverse process, the noisy image is gradually denoised back to a clean state,
\begin{equation}
\label{eq2}
x_{t-1}=\sqrt{\alpha_{t-1}} \frac{x_{t}-\sqrt{1-\alpha_{t}} \epsilon_{\theta}\left(x_{t}, t\right)}{\sqrt{\alpha_{t}}}+\sqrt{1-\alpha_{t-1}} \epsilon_{t},
\end{equation}
where $\alpha_{t-1}=\frac{\bar{\alpha}_{t-1}}{\bar{\alpha}_t}$ for $t = T,..., 1$. $\epsilon_{\theta}\left(x_{t}, t\right)$ is the predicted noise by the denoising neural network parameterized by $\theta$, typically implemented as a U-Net.

\subsection{Architecture}
Recent methods \cite{wang2023dire, chu2024fire} propose to use reconstruction error as a feature for detecting diffusion-generated images. However, these methods nearly fail in the detection of diffusion-based inpainting images (see Section Experiments). We analyze several limitations of these methods: (1) The reconstruction and the detection phases are completely independent, lacking alignment in the latent space; (2) They only focus on global features of the image, neglecting local regions. To this end, we propose End4, which couples the reconstruction and detection phases through an updated diffusion reconstruction model and emphasizes local features using a multi-scale pyramid-like attention module.

The pipeline of our method is shown in Figure \ref{pipeline}. First, we use the forward diffusion process proposed in \cite{ho2020denoising} to corrupt the original image $x_{0}$ into $x_{t}$ at a random time step $t$ using Equation \ref{eq1}. As the time step increases, the original image $x_{0}$ gradually loses its discriminative features and approaches an isotropic Gaussian distribution. Then $\epsilon_{\theta}\left(x_{t}, t\right)$ is a function approximator intended to predict the noise $\epsilon$ from $x_{t}$ and $t$, implemented using a U-Net \cite{dhariwal2021diffusion, ronneberger2015u}-like architecture based on PixelCNN \cite{salimans2017pixelcnn++}, ResNet \cite{he2016deep}, and Transformer \cite{vaswani2017attention}.  Therefore, the reconstruction image can be obtained iteratively from Equation \ref{eq2}. Most existing methods utilize DDIM \cite{song2020denoising} to accelerate the iterative process without the Markov hypothesis. Each iteration of the denoising process in the diffusion model corresponds to a round of network inference, which requires substantial computational resources and time. This poses a significant challenge for real-time inference. To address this, we adopt a direct reconstruction method, referred to as \textit{one-step denoising}, as an alternative to the iterative approach. Specifically, at any time step $t$, after the diffusion model predicts the noise $\epsilon_\theta(x_t, t)$ of $x_{t}$ by a single inference (one-step), direct recovery is always valid \cite{ho2020denoising}, as follows:
\begin{equation}
\label{eq3}
\hat{x}_{0}=\frac{1}{\sqrt{\bar{\alpha}_{t}}}\left( x_{t}-\sqrt{1-\bar{\alpha}_{t}} \epsilon_{\theta}\left( x_{t}, t \right) \right),
\end{equation}
where $\hat{x}_{0}$ refers to the reconstruction via \textit{one-step denoising}. This direct prediction is $t$ times faster than iterative prediction, resulting in significant savings of computational resources and inference time.

Next, we feed the original image $x_{0}$ and the reconstructed image $\hat{x}_{0}$ into the Scale-aware Pyramid-like Fusion Module (SPFM). Finally, we input the fused feature map into a binary classifier for the final prediction:
\begin{equation}
\label{eq4}
y^{\prime}=\theta_{cls}\left(\mathcal{M}\left(x_{0}, \hat{x}_{0}\right)\right),
\end{equation}
where $\mathcal{M}(\cdot)$ represents the SPFM. $y^{\prime}$ is the predicted label. $\theta_{cls}$ represents the binary classifier. 

\subsection{Scale-aware Pyramid-like Fusion Module}
Most existing methods directly concatenate the features of the original and the reconstructed images after passing through a backbone (e.g. ResNet50), neglecting the interaction between them. To better capture their global and local features and achieve fusion interaction, we propose a novel Scale-aware Pyramid-like Fusion Module (SPFM), as shown in Figure \ref{fig:SPFM}. First, we pass the original image ${x}_{0}$ and the reconstructed image $\hat{x}_{0}$ through a ResNet50 \cite{he2016deep} network to obtain features $x$ and $\hat{x}$, respectively. Then, we serially employ two scale-aware pyramid-like layers to obtain features with different scale sensitivities. Specifically, after performing attention operations via a transformer block, we acquire local features through several convolutional blocks and enhance positional information. Both pyramid-like structures are identical, differing in the size of the attention mask and the dilation convolution, which are $2*size$ and $size$, respectively. The $size$ is set to 2 and 4, respectively. Through the pyramid-like structure, we obtain feature vectors with different scale sensitivities. Next, we stack them as follows,
\begin{equation}
\label{eq5}
x_{m} = Stack(L_1(x), L_2(L_1(x))),
\end{equation}
\begin{equation}
\label{eq6}
\hat{x}_{m} = Stack(L_1(\hat{x}), L_2(L_1(\hat{x}))),
\end{equation}
where $Stack(\cdot)$ denotes the stacking operation. $L_1$ and $L_2$ represent the scale-aware pyramid-like layer 1 and layer 2, respectively.

Then, self-attention is performed to obtain attention features with different scale sensitivities,
\begin{equation}
\label{eq7}
x_{s} = Self(x_{m}),
\end{equation}
\begin{equation}
\label{eq8}
\hat{x}_{s} = Self(\hat{x}_{m}),
\end{equation}
where $Self(\cdot)$ denotes the self-attention operation.

To learn semantic correlations between the features of the original image and the reconstructed image, we employ the multi-head cross-attention mechanism from Transformer \cite{vaswani2017attention}. Finally, both features are concatenated to obtain the final feature $x_\mathrm{SPFM}$,
\begin{equation}
\label{eq9}
x_\mathrm{SPFM} = Concat(MCA(x_{s}, \hat{x}_{s}), MCA(\hat{x}_{s}, x_{s})),
\end{equation}
where $Concat(\cdot)$ denotes the concatenation operation, and $MCA(Q, KV)$ represents multi-head cross-attention. $x_\mathrm{SPFM}$ is followed by an FC layer for classification.

\subsection{Loss Function}
We define $x_{0}$ as the original image, which is either nature ($y=0$) or inpainting ($y=1$). We jointly train the reconstruction and classification branches in the end-to-end paradigm. The diffusion model learns the entire distribution of inpainting samples (\textit{i.e.}, $y=1$) by minimizing the following \textbf{noise loss}, which can be formulated as:
\begin{equation}
\label{eq10}
\mathcal{L}_{noise}=\frac{y\|\epsilon_{t}-\epsilon_\theta(x_{t}, t)\|^2}{2},
\end{equation}
where $\epsilon_{t}$ is the added noise and $\epsilon_\theta(x_{t}, t)$ is the predicted noise. We then employ the cross-entropy loss to train the binary classifier:
\begin{equation}
\label{eq11}
\mathcal{L}_{ce}=-\left[\left(y \log \left(y^{\prime}\right)+(1-y) \log \left(1-y^{\prime}\right)\right)\right],
\end{equation}
where $y$ and $y^{\prime}$ refer to the ground truth and model prediction. Finally, the \textbf{overall loss objective} of our proposed method minimizes the combination of the noise loss and the cross-entropy loss, \textit{i.e.},
\begin{equation}
\label{eq12}
\mathcal{L} = \mathcal{L}_{noise} + \mathcal{L}_{ce}.
\end{equation}
\section{Experiments}
\label{sec:exprs}

\begin{figure*}[!t]
\centering
\includegraphics[width=5.2in]{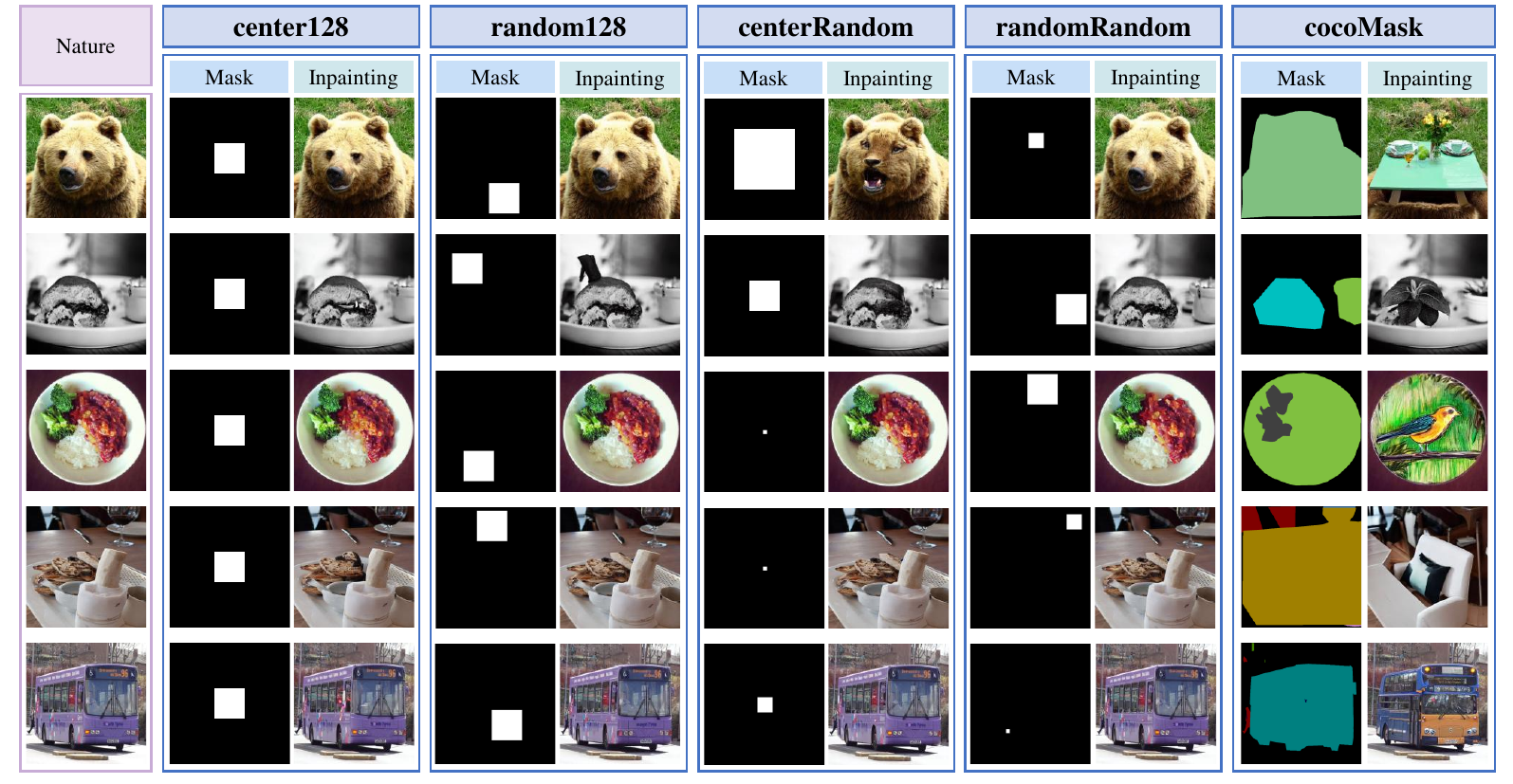}
\caption{Samples of the InpaintingForensics dataset from SD v1.5 Inpainting.}
\vspace{-0.3em}
\label{fig:dataset}
\end{figure*}

\subsection{InpaintingForensics: A New Dataset for Evaluating Diffusion-Based Inpainting Detectors}

To better evaluate the performance of the diffusion-based inpainting detectors, we collect a dataset called InpaintingForensics, which consists of five subsets for comprehensive experiments. The images are sourced from the Common Objects in Context (COCO) dataset \cite{lin2014microsoft}. The dataset consists of a train, validation, and test split and feature annotations such as image descriptions and segmentation masks. 
We utilize 5000 images from the validation set, ensuring diversity in resolution, content, and visual quality. All images are resized to 512×512. We create the following five subsets based on variations in the masks' shape, size, and position:

(1) \textbf{center128:} A mask of size 128×128 is selected at the center of the image. The text prompt for inpainting is “a photo”.

(2) \textbf{random128:} A mask of size 128×128 is selected at a random position in the image. The text prompt for inpainting is “a photo”.

(3) \textbf{centerRandom:} A mask of random size (including 16×16, 32×32, 64×64, 128×128, 256×256) is selected at the center of the image. The text prompt for inpainting is “a photo”.

(4) \textbf{randomRandom:} A mask of random size (same as above) is selected at a random position in the image. The text prompt for inpainting is “a photo”.

(5) \textbf{cocoMask:} Segmentation annotations from the COCO dataset \cite{lin2014microsoft} are used as masks. For each image, the category annotations from the COCO dataset \cite{lin2014microsoft} are obtained. If the category annotations for an image with $n$ categories are $\left \{k_1, k_2, …, k_n \right\} (n>0)$, a minimum of $(n, m)$ categories are randomly replaced from the total category pool (which contains 80 categories), where $m$ is set to 5. The replaced categories are the text prompt for inpainting.

Once the inpainting masks and text prompts are obtained, a diffusion-based inpainting model is employed to synthesize the images. We use SD v1.5 Inpainting\footnote{runwayml/stable-diffusion-inpainting}, SD v2 Inpainting\footnote{stabilityai/stable-diffusion-2-inpainting}, and BrushNet\footnote{https://github.com/TencentARC/BrushNet}\cite{ju2024brushnet} as the inpainting model with DDIM \cite{song2020denoising} scheduler applied for 20 steps. The resulting 5000 images in each subset are divided into two partitions: a training set consisting of 4000 images and a testing set consisting of 1000 images. The dataset consists of approximately 75k fake images and includes nature images, mask images, and inpainting images. Samples of the InpaintingForensics dataset are shown in Figure \ref{fig:dataset}, including nature, mask and inpainting images with five different mask regions.

\begin{table*}[!t]
\centering
\vspace{0.3em}
\begin{tabular}{lccccc}
\toprule
\multirow{2}{*}[-0.6ex]{Method} & \multicolumn{5}{c}{Testing Subset}                                                                               \\ \cmidrule(l){2-6}  
                        & center128 & random128 & centerRandom & randomRandom & cocoMask \\ \midrule

F3Net \cite{qian2020thinking}                   & 51.95/51.33          & 50.25/50.41          & 50.25/49.44          & 50.20/49.99          & 48.51/46.27          \\
DIRE \cite{wang2023dire}                   & 50.20/49.99          & 49.95/50.29          & 49.70/48.94          & 50.00/51.01          & 53.18/54.47          \\
UFD \cite{ojha2023towards}                    & 50.25/50.05          & 49.40/49.37          & 49.05/49.66          & 49.65/49.55          & 47.80/47.51          \\
NPR \cite{tan2024rethinking}                    & 65.90/71.07          & 66.35/71.39          & 66.10/71.38          & 66.55/71.63          & 73.62/82.09          \\
FIRE \cite{chu2024fire}                   & 50.00/51.52          & 50.00/49.45          & 50.00/50.10          & 50.00/48.87          & 51.23/50.73          \\
Ours                    & \textbf{87.80/95.00} & \textbf{87.65/94.82} & \textbf{87.45/94.72} & \textbf{88.00/94.81} & \textbf{89.96/96.32} \\ \bottomrule
\end{tabular}
\caption{\textbf{Comparison of our method and other competitive state-of-the-art detectors.} We train models on the cocoMask subset of the InpaintingForensics dataset and evaluate on all five subsets (all forged data is generated using the SD v1.5 inpainting). All the models are retrained with the official codes. We report ACC (\%) and AUC (\%) (ACC/AUC in the table).}
\label{tab:comparison}
\end{table*}

\subsection{Experimental Setup}
\label{setup}
\textbf{Datasets.} We use the self-collected InpaintingForensics dataset, which contains three forgery methods and five mask settings. To evaluate the effectiveness and generalizability of our method, we train on the cocoMask subset and test on all five subsets.

\textbf{Baselines.} We compare our method with a set of state-of-the-art detectors. All methods use the official open-source code for training and inference on our experimental datasets. 1) \textbf{F3Net} \cite{qian2020thinking} proposes that the frequency information of images is essential for fake image detection. 2) \textbf{DIRE} \cite{wang2023dire} explores the image-level DDIM reconstruction error as a detection clue. 3) \textbf{UFD} \cite{ojha2023towards} trains a linear classification head based on CLIP \cite{radford2021learning} embeddings of real and fake images. 4) \textbf{NPR} \cite{tan2024rethinking} introduces neighborhood pixel relationships to capture up-sampling artifacts. 5) \textbf{FIRE} \cite{chu2024fire} utilizes frequency-guided reconstruction error as a detection clue.

\textbf{Implementation details.} For data preprocessing, we apply a series of random augmentations, including horizontal flip, color jitter, and grayscale. All images are resized to 256×256. For the denoising network, we employ a U-Net \cite{ronneberger2015u} architecture to predict noise, utilizing one-step denoising with a random time step $t\in(0,1000]$. We use the ResNet50 \cite{he2016deep} network for the feature extraction. During training, the batch size is set to 8, and the Adam optimizer \cite{kingma2014adam} is employed with an initial learning rate of 1\textit{e}-4. All experiments are conducted on a machine equipped with NVIDIA A100 GPUs. We adopt two widely used metrics for image generation detection: accuracy (ACC) and Area Under the ROC Curve (AUC) to evaluate the effectiveness of models.

\subsection{Comparison to Existing Detectors}

In this section, we compare our method with several state-of-the-art image generation detectors. We train on the cocoMask subset from SD v1.5 inpainting model of the InpaintingForensics dataset. To validate the models’ generalization capability across varying masks, we evaluate on five subsets with different masks. The experimental results are shown in Table \ref{tab:comparison}. It is evident that many previous methods, which are highly effective in deepfake detection \cite{qian2020thinking} and diffusion-generated image detection \cite{wang2023dire, ojha2023towards, chu2024fire}, fail in diffusion-based inpainting detection, achieving only about 50\% accuracy. The detection performance of NPR \cite{tan2024rethinking} is slightly better but does not exceed 75\%. We analyze that previous methods focus on the global detection of images, lacking fine-grained local detection, which is not suitable for the task of local forgery inpainting detection. In contrast, our method achieves accuracy of over 87\%, significantly outperforming other methods.

\subsection{Ablation Study}
\subsubsection{Influence of SPFM}
In this experiment, we investigate the impact of different fusion methods (between the original image and the reconstructed image) on model detection performance. Specifically, we use the absolute difference, concatenation, and SPFM proposed in the paper for fusing the original and reconstructed images. The results are presented in Table \ref{tab:SPFM}.
When detecting the cocoMask subset, the detection accuracy for the absolute difference and concatenation methods are 51.79\% and 71.62\%, respectively. In contrast, when using SPFM for fusion, the accuracy reaches 89.96\%. The experimental results indicate that the SPFM achieves a higher accuracy compared to the other methods. This finding suggests that SPFM is able to capture more of the local information from the images. In comparison, other methods result in the loss of critical details.

\begin{table}[!t]
\centering
\label{tab:SPFM}
\vspace{0.3em}
\scalebox{0.9}{
\begin{tabular}{@{}cccccc@{}}
\toprule
\multirow{2}{*}[-0.6ex]{Method} & \multicolumn{5}{c}{Testing Subset}                                                 \\ \cmidrule(l){2-6} 
                        & c128      & r128      & cR   & rR   & coco       \\ \midrule
Absolute   Difference   & 51.15          & 50.80          & 50.70          & 50.00          & 51.79          \\
Concatenation           & 67.55          & 67.10          & 67.10          & 65.85          & 71.62          \\
SPFM (Ours)                   & \textbf{87.80} & \textbf{87.65} & \textbf{87.45} & \textbf{88.00} & \textbf{89.96} \\ \bottomrule
\end{tabular}
}
\caption{Ablation study on different fusion methods, measured in ACC(\%). We train on cocoMask (coco) and test on center128 (c128), random128 (r128), centerRandom (cR), and randomRandom (rR).}
\label{tab:SPFM}
\end{table}

\subsubsection{Influence of Loss Terms}
In this section, we conduct ablation experiments on the target of $\mathcal{L}_{noise}$. The results are shown in Table \ref{tab:loss}. We first ablate $\mathcal{L}_{noise}$, which means the reconstruction model lacks guidance from the loss function. The results show that the model’s performance drops under this setting. Next, we compute $\mathcal{L}_{noise}$ for all images. The model’s performance decreases more significantly, suggesting that indiscriminate loss guidance for both nature and inpainting images is ineffective. Then, we apply $\mathcal{L}_{noise}$ to nature images and inpainting images, respectively. The results show that providing $\mathcal{L}_{noise}$ only for inpainting images yields better performance. This aligns with the principle of diffusion generation detection based on reconstruction: generated images and their reconstructed images are more similar than real images and their reconstructions. These analyses indicate that selecting an appropriate target for $\mathcal{L}_{noise}$ is crucial for the performance of the model.

\begin{table}[!t]
\centering
\vspace{0.3em}
\scalebox{0.9}{
\begin{tabular}{@{}ccccccc@{}}
\toprule
\multicolumn{2}{c}{$\mathcal{L}_{noise}$} & \multicolumn{5}{c}{Testing   Subset}                                               \\ \cmidrule(r){1-2} \cmidrule(l){3-7}
nature     & inpainting    & c128      & r128      & cR   & rR   & coco       \\ \midrule
$\times$           & $\times$              & 75.25          & 74.80          & 75.10          & 75.30          & 77.36          \\
\checkmark          & \checkmark             & 65.20          & 64.80          & 65.30          & 64.80          & 71.98          \\
\checkmark          & $\times$              & 76.90          & 77.35          & 78.15          & 77.00          & 80.89          \\
$\times$           & \checkmark             & \textbf{87.80} & \textbf{87.65} & \textbf{87.45} & \textbf{88.00} & \textbf{89.96} \\ \bottomrule
\end{tabular}
}
\caption{Ablation study on loss terms, measured in ACC(\%).}
\label{tab:loss}
\end{table}

\begin{figure*}[!t]
\centering
\includegraphics[width=5.5in]{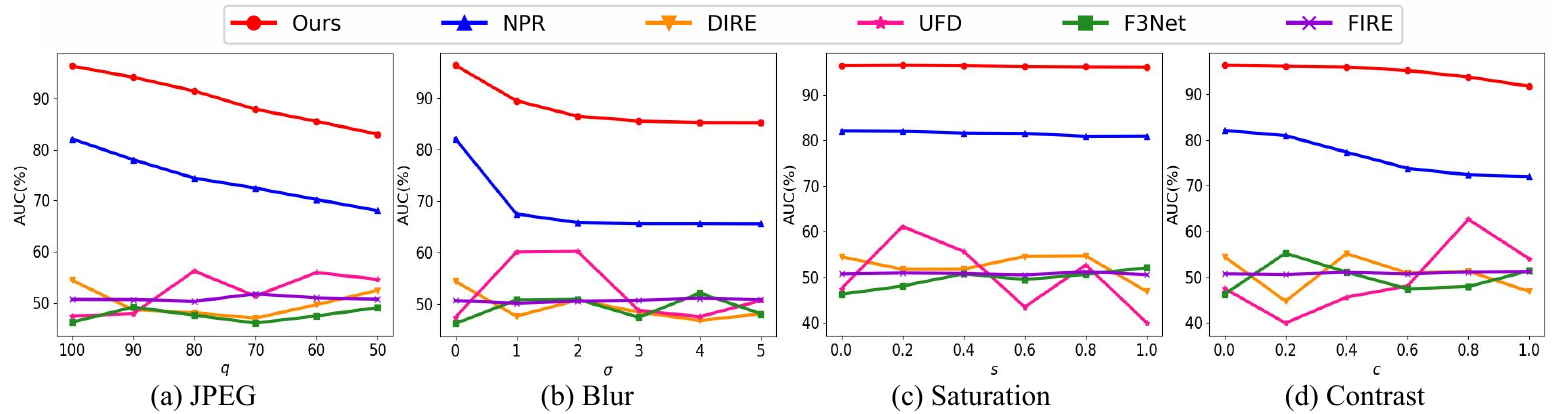}
\caption{Robustness of baselines and ours to various unseen perturbations, measured in AUC(\%).}
\label{fig:robust}
\end{figure*}

\begin{figure*}[!t]
\centering
\includegraphics[width=5.5in]{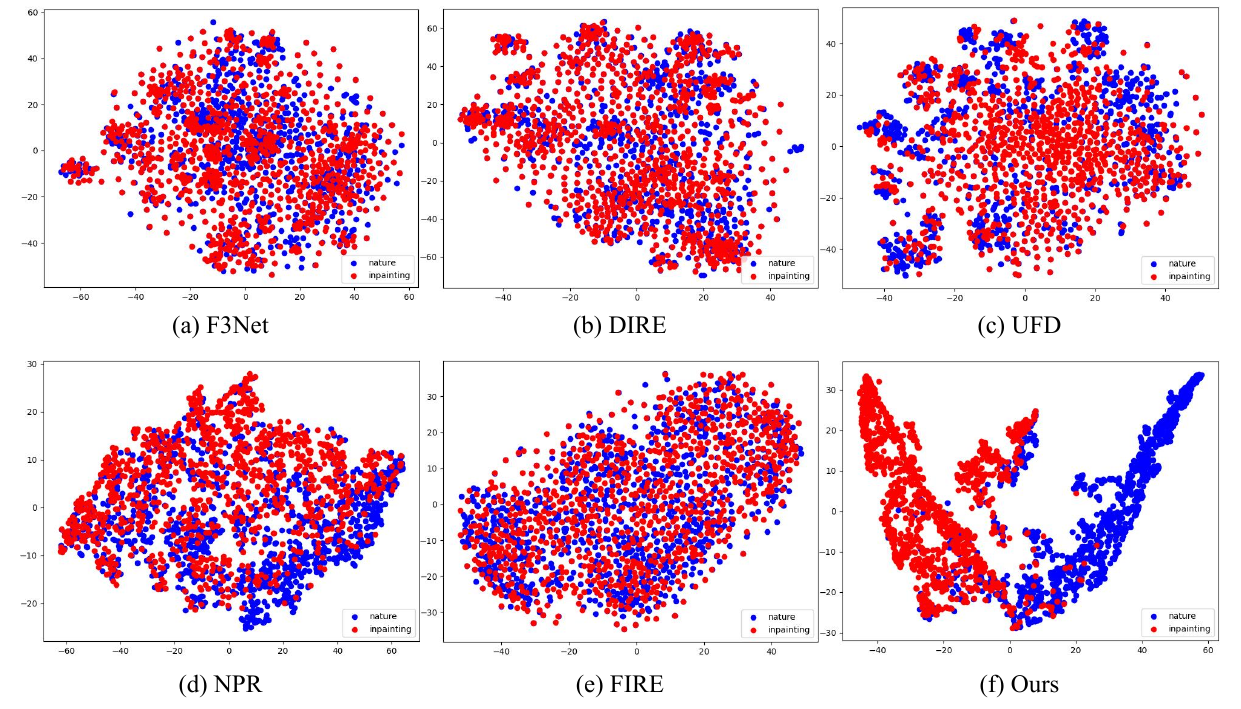}
\caption{Feature distribution of models via t-SNE \cite{van2008visualizing}.}
\label{fig:tsne}
\end{figure*}

\subsection{Robustness to Unseen Perturbations}
In real-world scenarios, the images to be detected are often post-processed, such as through quality compression or Gaussian blur. In this section, we evaluate the robustness of our method against various perturbations. Following previous works \cite{chu2024fire, wang2020cnn}, we use the cocoMask subset and apply JPEG compression (quality factor $q$), Gaussian blur (standard deviation $\sigma$), saturation (saturation factor $s$), and contrast (contrast factor $c$). The results are shown in Figure \ref{fig:robust}. In the figure, we report the performance of several baselines and our method under varying levels of image perturbations. We observe that at each level of perturbation, our method maintains the best performance, outperforming the baselines. This indicates that our method is robust enough to handle real-world scenarios.

\subsection{Visual Analysis}
We utilize the t-SNE \cite{van2008visualizing} visualization to impressively illustrate the feature representations extracted from the last layer of our proposed model and the baseline methods, as shown in Figure \ref{fig:tsne}. Each model is trained and evaluated on the cocoMask subset of the InpaintingForensics dataset. Although several models exhibit overlapping distributions of nature and inpainting images in the feature space, our model shows a smaller degree of overlap. The nature and inpainting images from the other five models are nearly indistinguishable in the feature space, whereas our model displays a distinct separation, indicating its better detection capability.

\section{Conclusion}
\label{sec:Conclusions}
In this paper, we focus on building a general detector for discriminating diffusion-based inpainting-generated images. We find that previous generated-image detectors exhibit limited performance when detecting diffusion-based inpainting images. To address this issue, we propose an end-to-end denoising diffusion-based detection method with a scale-aware pyramid-like fusion module. Additionally, we create a new dataset, InpaintingForensics, which includes inpainting images generated with five different masked regions. Extensive experiments demonstrate that our method outperforms state-of-the-art baselines, achieving superior performance both on standard datasets and under challenging conditions with perturbed images. We hope that our work can serve as a solid baseline for diffusion-based inpainting detection.

\newpage
\bibliography{aaai2026}

\end{document}